\pdfoutput=1
\documentclass[journal]{IEEEtran}
\usepackage[colorlinks=true,linkcolor=red,citecolor=blue,urlcolor=blue]{hyperref}
\usepackage[cmex10]{amsmath}
\usepackage{amssymb}
\usepackage{graphicx}
\usepackage{tabularray}
\usepackage{bm}
\usepackage{eso-pic}
\usepackage[style=ieee,backend=biber,sorting=none]{biblatex}
\UseTblrLibrary{booktabs}
\DeclareFieldFormat{bibliography}{\setlength{\baselineskip}{1.0\baselineskip}#1}

\begin{document}

\AddToShipoutPictureBG*{%
  \AtPageLowerLeft{%
    \put(\LenToUnit{0.5\paperwidth},\LenToUnit{0.34in}){%
      \makebox(0,0)[c]{\parbox{0.9\textwidth}{\centering\fontsize{6.5}{7.5}\selectfont
        \textcopyright~2026 IEEE. Personal use of this material is permitted. Permission from IEEE must be obtained for all other uses, in any current or future media, including reprinting/republishing this material for advertising or promotional purposes, creating new collective works, for resale or redistribution to servers or lists, or reuse of any copyrighted component of this work in other works.}}}%
  }%
}

\title{ORDR-DETR: Objectness-Guided Query Ranking and Detail Refinement for Fine-Grained\\Object Detection}

\author{Wentao~Li,
  Danpei~Zhao,~\IEEEmembership{Senior Member,~IEEE,}
  Bo~Yuan,
  Tian~Li,
  and~Zhiguo~Jiang,~
  \IEEEmembership{Member,~IEEE}%
  \thanks{This work was supported by the National Natural Science Foundation of China under Grant 62271018. (Corresponding author: Danpei Zhao.)}%
  \thanks{Wentao Li, Danpei Zhao, Bo Yuan, Tian Li, and Zhiguo Jiang are with the Department of Aerospace Intelligent Science and Technology, School of Astronautics, Beihang University, Beijing 102206, China, and the Key Laboratory of Spacecraft Design Optimization and Dynamic Simulation Technology, Ministry of Education, Beihang University, Beijing 102206, China. Danpei Zhao, Bo Yuan, Tian Li, and Zhiguo Jiang are also with the Tianmushan Laboratory, Beihang University, Hangzhou 311115, China. (e-mail: canoe@buaa.edu.cn; zhaodanpei@buaa.edu.cn; yuanbobuaa@buaa.edu.cn; lit@buaa.edu.cn; jiangzg@buaa.edu.cn).}}

\markboth{IEEE Geoscience and Remote Sensing Letters}%
{Li \MakeLowercase{\textit{et al.}}: ORDR-DETR}

\maketitle

%% ===================== ABSTRACT =====================
\begin{abstract}
DETR-based fine-grained object detection (FGOD) remains challenging because the shared classification score is used for both query ranking and fine-grained recognition, and decoder queries lack local details from low-level RoI features. To address these issues, we propose ORDR-DETR (Objectness-guided query Ranking and Detail Refinement DETR), which decouples query ranking from fine-grained recognition and enhances decoder queries with local detail refinement. First, we propose OGQR, which splits the head into a ranking branch and a fine-grained classification branch. The ranking branch is trained with one-to-one supervision for quality-aware ranking, while the classification branch uses one-to-many supervision for fine-grained semantics. We further propose DQR, which injects rotation-aligned RoI features into late decoder queries to improve fine-grained discrimination. On MAR20, ShipRSImageNet, and FAIR1M-v2.0, ORDR-DETR achieves state-of-the-art AP$_{50}$ of 85.80, 76.30, and 50.97, respectively, surpassing strong two-stage FGOD methods with an end-to-end DETR pipeline. Code is available at \url{https://github.com/AIRVIC-Lab/ORDR-DETR}.
\end{abstract}


\IEEEpeerreviewmaketitle

% ---- Figure 1 ----
\begin{figure*}[!t]
  \centering
  \includegraphics[width=0.9\textwidth]{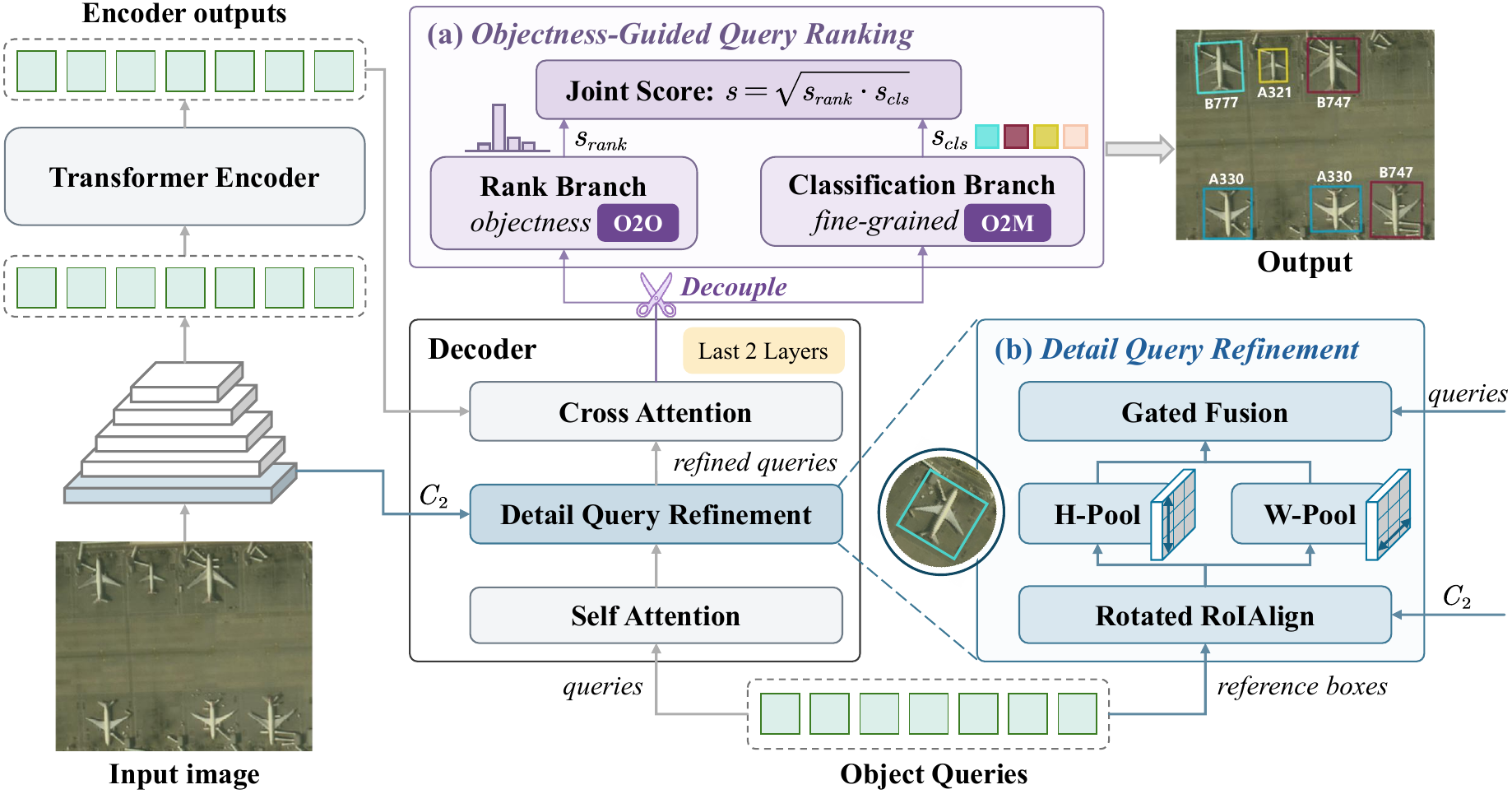}
  \caption{Overview of ORDR-DETR. (a) OGQR decouples query ranking and fine-grained classification. (b) DQR injects rotation-aligned detail features into late decoder queries.}
  \label{fig:overview}
\end{figure*}

%% ===================== INTRODUCTION =====================
\section{Introduction}

\IEEEPARstart{F}{ine-grained} object detection (FGOD) in remote sensing requires not only accurate object localization but also discriminative recognition of fine-grained classes, for example, distinguishing a Boeing 737 from an Airbus A320 rather than simply detecting an airplane. Compared with general object detection in remote sensing, FGOD is more challenging because large intra-class variance and ambiguous inter-class differences often cause semantic confusion. As aerial objects often appear at arbitrary orientations, existing FGOD methods are mainly built on oriented detectors. Most of them follow the two-stage paradigm and separate proposal generation from fine-grained recognition. Recent two-stage FGOD methods improve FGOD by refining RoI features or decoupling localization and classification representations~\cite{pcldet,sfrnet,petdet,drnet}. While effective, these methods rely on dense proposals and NMS post-processing.

In general object detection, DETR and its variants~\cite{detr,deform,dino,rtdetr} formulate detection as set prediction and enable end-to-end detection without NMS. Query-based oriented detectors have also developed rapidly in remote sensing~\cite{arsdetr,oformer,rhino,o2rtdetr}. However, DETR methods designed for FGOD remain limited. Existing DETR-based FGOD methods enhance DETR through rotation-aligned attention and multiscale query fusion~\cite{qedetr}. However, the classification score still serves both query ranking and fine-grained recognition. These two roles demand different supervision, and this conflict remains unaddressed in DETR-based FGOD.

Specifically, when applied to FGOD, one-to-one matching for query ranking can conflict with fine-grained semantic learning. A query may localize an object well but still be treated as a negative sample. This weakens the discriminative learning needed by FGOD. Semantic ambiguity among similar fine-grained classes can also make the matching unstable. For these two roles, query ranking needs a sparse and quality-aware signal, while fine-grained recognition benefits from denser supervision on multiple queries around the same instance. A single classification branch is not well suited to both roles. DETR-style methods also lack the local detail that two-stage methods obtain from low-level backbone features through RoI pooling. As a result, queries may miss subtle cues that distinguish similar fine-grained classes.

In this paper, we propose ORDR-DETR, a query-based FGOD framework that addresses these two issues. We first introduce Objectness-Guided Query Ranking (OGQR), which splits the detection head into a class-agnostic ranking branch and a fine-grained classification branch. The ranking branch uses one-to-one assignment to learn a quality-aware objectness score, and the classification branch uses one-to-many assignment. This separation distinguishes OGQR from quality-aware classification~\cite{vfl}, which keeps localization quality inside the class score and uses it for both ranking and recognition. We then introduce Detail Query Refinement (DQR), which extracts rotation-aligned RoI features from a high-resolution feature map and injects them into late decoder queries through gated residual fusion. Unlike two-stage FGOD methods~\cite{petdet,pcldet} that feed such features into a separate refinement head, DQR updates the query state directly.

Experiments on MAR20~\cite{mar20}, ShipRSImageNet~\cite{shiprs}, and FAIR1M-v2.0~\cite{fair1m} show that ORDR-DETR consistently outperforms existing methods. Compared with current state-of-the-art methods, it brings AP$_{50}$ improvements of +0.34, +1.00, and +1.63, respectively. Our contributions are as follows:
\begin{enumerate}
  \item We propose ORDR-DETR, a DETR-based framework for FGOD that addresses the mismatch between query ranking and fine-grained recognition, improving both ranking quality and fine-grained discrimination.
  \item We propose two collaborative modules, where OGQR decouples ranking and classification with role-suited supervision, and DQR enhances queries with rotation-aligned local details for better fine-grained recognition.
  \item ORDR-DETR achieves state-of-the-art performance on MAR20, ShipRSImageNet, and FAIR1M-v2.0, outperforming strong two-stage FGOD detectors within an end-to-end pipeline.
\end{enumerate}

%% ===================== METHOD =====================
\section{Method}

\subsection{Overview}

The overall framework of ORDR-DETR is shown in Fig.~\ref{fig:overview}. Built on RT-DETR-O~\cite{rtdetr}, ORDR-DETR addresses two limitations of DETR-based FGOD: the mismatch between query ranking and fine-grained recognition, and the lack of local detail in decoder queries. OGQR separates query ranking from fine-grained classification with different supervision. DQR extracts rotation-aligned detail features from the $C_2$ feature map and injects them into late decoder queries.

\subsection{Objectness-Guided Query Ranking}

In standard DETR, the classification scores are shared between query ranking and class prediction, which creates conflicting supervision demands in FGOD. OGQR splits the shared classification branch into a ranking branch and a classification branch.

The ranking branch produces a class-agnostic objectness score from the query feature $\bm{q} \in \mathbb{R}^{d}$:
\begin{equation}
  s_{\mathrm{rank}} = \sigma(\bm{W}_r \bm{q}) \label{eq:rank}
\end{equation}
where $\bm{W}_r \in \mathbb{R}^{1 \times d}$. We use one-to-one assignment for the ranking branch and optimize it with Varifocal Loss~\cite{vfl}. In this way, the ranking score is trained to reflect both objectness and localization quality.

We use a separate classification branch to predict the joint fine-grained classification scores:
\begin{equation}
  \bm{s}_{\mathrm{cls}} = \sigma(\bm{W}_c \bm{q}) \label{eq:cls}
\end{equation}
where $\bm{W}_c \in \mathbb{R}^{C \times d}$ and $C$ is the number of fine-grained classes. We use one-to-many assignment for this branch and set the positive threshold to 0.5. This provides more positive queries for each ground-truth instance and helps fine-grained semantic learning. The regression branch follows the baseline.

At inference, the final score for class $c$ is the geometric mean of the two branches:
\begin{equation}
  s_{c} = \sqrt{s_{\mathrm{rank}} \cdot s_{\mathrm{cls},c}} \label{eq:fusion}
\end{equation}

During training, OGQR is also compatible with contrastive denoising training~\cite{dino}. For denoising queries, the ranking branch uses binary objectness labels, while the classification branch receives fine-grained class labels. We use 100 denoising queries by default.

\subsection{Detail Query Refinement}

DETR-style methods do not explicitly extract local RoI features, so decoder queries may lack the local details needed for fine-grained recognition. DQR addresses this issue by extracting rotation-aligned detail features from the $C_2$ feature map and injecting them into late decoder queries.

For each query with reference box $\bm{b}$, we extract a $K \times K$ rotation-aligned RoI feature $\bm{F} \in \mathbb{R}^{D' \times K \times K}$ from the $C_2$ feature map, where $K$ is the RoI size and is set to 5 by default. Before RoI extraction, the $C_2$ feature map is projected to 128 dimensions by a $1\times1$ convolution. Since many oriented objects are elongated, we pool $\bm{F}$ along the two box axes separately:
\begin{align}
  \bm{h} & = \mathrm{AvgPool}_{K \times 1}(\bm{F}) \nonumber       \\
  \bm{w} & = \mathrm{AvgPool}_{1 \times K}(\bm{F}) \label{eq:pool}
\end{align}

The two pooled features are projected and concatenated to form a detail token $\bm{d}$. We predict a channel-wise gate $\bm{g}$ from the concatenated feature $[\bm{q}; \bm{d}]$ and use it to control how much detail information is injected into the query:
\begin{align}
  \bm{g}  & = \sigma\!\left(\bm{W}_2 \, \phi\!\left(\bm{W}_1 [\bm{q}; \bm{d}]\right)\right) \nonumber \\
  \bm{q}' & = \mathrm{LN}\!\left(\bm{q} + \bm{g} \odot \bm{d}\right) \label{eq:gate}
\end{align}
where $\bm{q}$ is the input query, $\bm{g}$ is a channel-wise gate, $[\cdot;\cdot]$ denotes concatenation, $\odot$ denotes element-wise multiplication, and $\phi$ denotes the SiLU activation.

DQR is inserted after self-attention in the last two decoder layers. At this point, each query has aggregated context from other queries and carries clearer semantics. Injecting local detail here lets the refined query interact with image features more effectively in the following cross-attention. We use DQR only in late layers because the reference boxes are more reliable there, leading to more accurate rotated RoI features.

\section{Experiments}

\subsection{Datasets and Evaluation Metrics}

We evaluate ORDR-DETR on MAR20~\cite{mar20}, ShipRSImageNet~\cite{shiprs}, and FAIR1M-v2.0~\cite{fair1m}. MAR20 contains 3,842 images with 22,341 instances across 20 aircraft classes, and many classes differ only in subtle structural details. ShipRSImageNet contains 3,435 images and 17,573 ship instances. We evaluate the level-3 setting with 50 fine-grained ship classes and use the validation split for testing because the test annotations are not public. FAIR1M-v2.0 is a large-scale remote-sensing FGOD benchmark with 37 fine-grained classes. Following prior works, we train the proposed model on the training and validation sets and submit predictions to the official evaluation server for testing.

We report AP$_{50}$ as the primary metric on all three datasets. For MAR20 and ShipRSImageNet, we also report mAP and AP$_{75}$. For FAIR1M-v2.0, we submit predictions to the official evaluation server, which reports AP$_{50}$ only.

\subsection{Implementation Details}

All experiments are implemented based on the MMRotate 1.x toolbox with 4 GPUs and a total batch size of 8. All compared methods are reproduced under a unified setting. Methods with the suffix ``-O'' denote our rotated adaptations of DETR baselines~\cite{deform,dino,rtdetr}. These variants use an angle-weighted L1 regression loss to stabilize matching and adopt the rotated deformable attention proposed in ARS-DETR~\cite{arsdetr} to align attention sampling. For all DETR-style methods, we use 300 queries by default. For models based on DINO, we use 100 contrastive denoising queries. We adopt R50-vd as the backbone for RT-DETR-O, O2-RT-DETR, and ORDR-DETR, following the default setting of RT-DETR. We use AdamW with learning rate $2 \times 10^{-4}$ and weight decay $10^{-4}$. For MAR20 and ShipRSImageNet, models are trained for 36 epochs with the learning rate reduced by a factor of 0.1 at epoch 30. For FAIR1M-v2.0, models are trained for 12 epochs with learning rate reduced at epoch 11.

For MAR20 and ShipRSImageNet, images are resized and padded to $800 \times 800$ and $1024 \times 1024$, respectively. For FAIR1M-v2.0, images are cropped into $1024 \times 1024$ patches with an overlap of 200. Random flipping is the only data augmentation. No NMS is applied during inference on MAR20 or ShipRSImageNet. For FAIR1M-v2.0, rotated NMS with an IoU threshold of 0.1 is used only for patch merging.

\subsection{Comparison With State-of-the-Art Methods}

% ---- Table I: MAR20 ----
\begin{table}[!t]
  \centering
  \caption{Comparison results with state-of-the-art methods on MAR20.}
  \label{tab:mar20}
  \begin{tblr}{
      colspec = {lcccc},
    }
    \toprule
    Method                          & Backbone & mAP            & AP$_{50}$      & AP$_{75}$      \\
    \midrule
    R$^3$Det~\cite{r3det}           & R50      & 35.56          & 64.24          & 33.69          \\
    S$^2$ANet~\cite{s2anet}         & R50      & 46.88          & 77.38          & 51.06          \\
    Oriented R-CNN~\cite{orcnn}     & R50      & 58.25          & 82.71          & 72.15          \\
    PCLDet~\cite{pcldet}            & R50      & 57.57          & 82.02          & 71.82          \\
    PETDet~\cite{petdet}            & R50      & 60.84          & 85.06          & 76.67          \\
    DRNet~\cite{drnet}              & R50      & 58.79          & 82.51          & 73.61          \\
    Deformable-DETR-O~\cite{deform} & R50      & 44.74          & 76.10          & 46.83          \\
    DINO-O~\cite{dino}              & R50      & 50.65          & 75.36          & 62.11          \\
    ARS-DETR~\cite{arsdetr}         & R50      & 51.13          & 76.28          & 63.47          \\
    OrientedFormer~\cite{oformer}   & R50      & 55.46          & 80.89          & 67.47          \\
    RHINO~\cite{rhino}              & R50      & 47.17          & 75.33          & 52.40          \\
    O2-RT-DETR~\cite{o2rtdetr}      & R50-vd   & 60.38          & 85.46          & 75.54          \\
    \midrule
    RT-DETR-O~\cite{rtdetr}         & R50-vd   & 61.58          & 84.23          & 77.18          \\
    \textbf{ORDR-DETR (Ours)}       & R50-vd   & \textbf{62.73} & \textbf{85.80} & \textbf{78.88} \\
    \bottomrule
  \end{tblr}
\end{table}

\begin{figure}[!t]
  \centering
  \makebox[\columnwidth][c]{%
    \makebox[\dimexpr0.285\columnwidth+2\fboxsep+2\fboxrule\relax][c]{\includegraphics[width=\dimexpr0.285\columnwidth+2\fboxsep+2\fboxrule\relax]{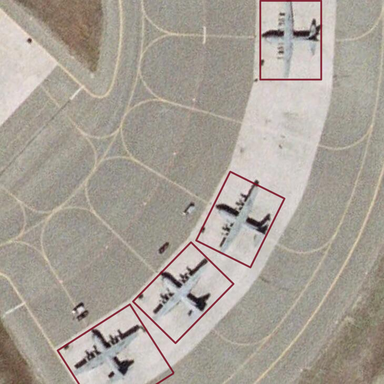}}%
    \hspace{2pt}%
    \makebox[\dimexpr0.285\columnwidth+2\fboxsep+2\fboxrule\relax][c]{\includegraphics[width=\dimexpr0.285\columnwidth+2\fboxsep+2\fboxrule\relax]{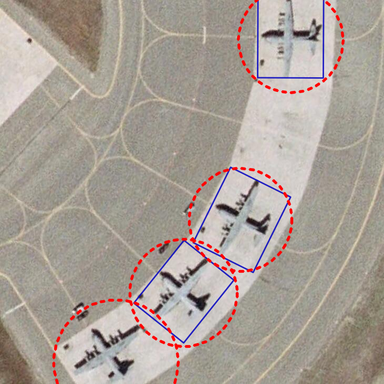}}%
    \hspace{2pt}%
    \makebox[\dimexpr0.285\columnwidth+2\fboxsep+2\fboxrule\relax][c]{\includegraphics[width=\dimexpr0.285\columnwidth+2\fboxsep+2\fboxrule\relax]{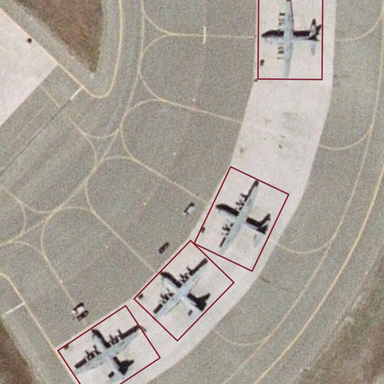}}%
  }\par

  \vspace{1pt}
  \makebox[\columnwidth][c]{%
    \makebox[\dimexpr0.285\columnwidth+2\fboxsep+2\fboxrule\relax][c]{\includegraphics[width=\dimexpr0.285\columnwidth+2\fboxsep+2\fboxrule\relax]{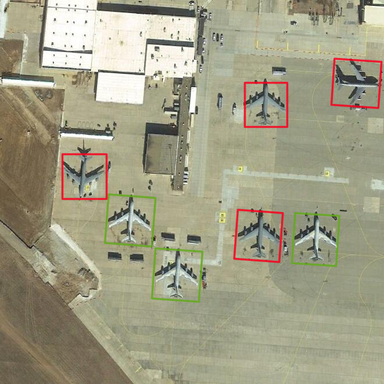}}%
    \hspace{2pt}%
    \makebox[\dimexpr0.285\columnwidth+2\fboxsep+2\fboxrule\relax][c]{\includegraphics[width=\dimexpr0.285\columnwidth+2\fboxsep+2\fboxrule\relax]{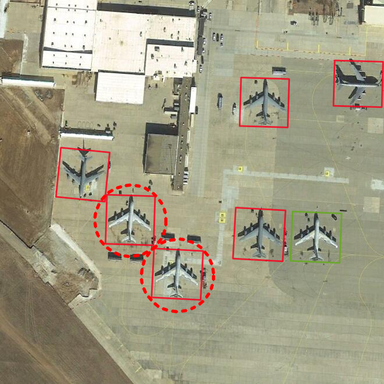}}%
    \hspace{2pt}%
    \makebox[\dimexpr0.285\columnwidth+2\fboxsep+2\fboxrule\relax][c]{\includegraphics[width=\dimexpr0.285\columnwidth+2\fboxsep+2\fboxrule\relax]{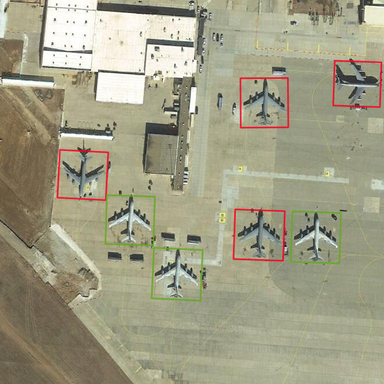}}%
  }\par

  \vspace{1pt}
  \makebox[\columnwidth][c]{%
    \makebox[\dimexpr0.285\columnwidth+2\fboxsep+2\fboxrule\relax][c]{\includegraphics[width=\dimexpr0.285\columnwidth+2\fboxsep+2\fboxrule\relax]{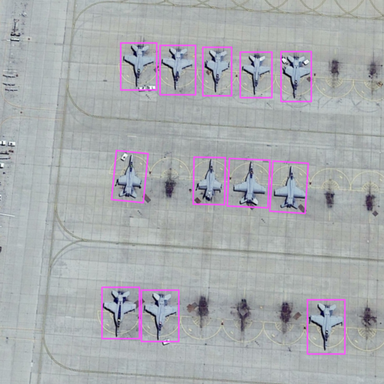}}%
    \hspace{2pt}%
    \makebox[\dimexpr0.285\columnwidth+2\fboxsep+2\fboxrule\relax][c]{\includegraphics[width=\dimexpr0.285\columnwidth+2\fboxsep+2\fboxrule\relax]{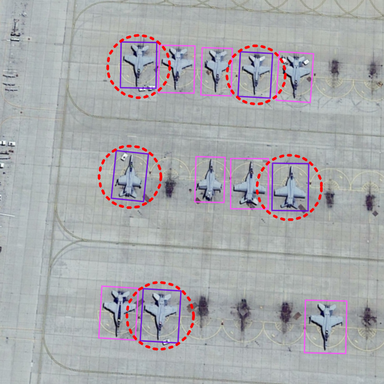}}%
    \hspace{2pt}%
    \makebox[\dimexpr0.285\columnwidth+2\fboxsep+2\fboxrule\relax][c]{\includegraphics[width=\dimexpr0.285\columnwidth+2\fboxsep+2\fboxrule\relax]{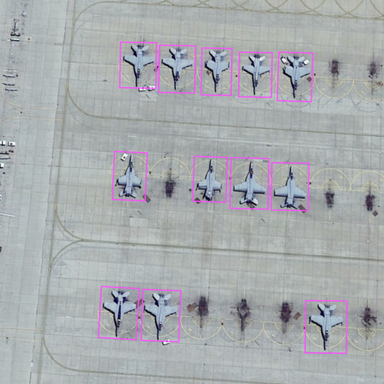}}%
  }\par
  \caption{Qualitative comparisons on MAR20. Columns show ground truth, RT-DETR-O, and ORDR-DETR. Only predictions with scores above 0.3 are shown. Different colors denote different fine-grained classes.}
  \label{fig:qualitative_result}
\end{figure}
On MAR20, fine-grained recognition is challenging because many aircraft classes look very similar. Table~\ref{tab:mar20} shows that ORDR-DETR achieves 62.73 mAP, 85.80 AP$_{50}$, and 78.88 AP$_{75}$. Compared with RT-DETR-O, it brings gains of +1.15 mAP, +1.57 AP$_{50}$, and +1.70 AP$_{75}$ at an additional cost of +1.34M parameters (42.80M to 44.15M) and +3.31G FLOPs (209.49G to 212.80G). It is also superior to other query-based oriented detectors and to representative two-stage FGOD methods, including PCLDet, PETDet, and DRNet. Qualitative examples in Fig.~\ref{fig:qualitative_result} show that the baseline still produces low-confidence predictions and semantic confusion. By contrast, ORDR-DETR predicts classes closer to the ground truth.

% ---- Table: ShipRSImageNet comparison ----
\begin{table}[!t]
  \centering
  \caption{Comparison results with state-of-the-art methods on ShipRSImageNet.}
  \label{tab:shiprs_sota}
  \begin{tblr}{
      colspec = {lcccc},
    }
    \toprule
    Method                          & Backbone & mAP            & AP$_{50}$      & AP$_{75}$      \\
    \midrule
    R$^3$Det~\cite{r3det}           & R50      & 19.51          & 32.82          & 20.90          \\
    S$^2$ANet~\cite{s2anet}         & R50      & 32.61          & 52.29          & 38.74          \\
    Oriented R-CNN~\cite{orcnn}     & R50      & 49.31          & 69.59          & 59.48          \\
    PCLDet~\cite{pcldet}            & R50      & 50.28          & 72.14          & 62.38          \\
    PETDet~\cite{petdet}            & R50      & 56.18          & 74.83          & 68.99          \\
    Deformable-DETR-O~\cite{deform} & R50      & 48.04          & 69.18          & 56.92          \\
    DINO-O~\cite{dino}              & R50      & 53.47          & 70.32          & 64.30          \\
    ARS-DETR~\cite{arsdetr}         & R50      & 43.98          & 63.45          & 51.59          \\
    OrientedFormer~\cite{oformer}   & R50      & 52.81          & 71.53          & 64.47          \\
    RHINO~\cite{rhino}              & R50      & 49.56          & 65.60          & 58.51          \\
    O2-RT-DETR~\cite{o2rtdetr}      & R50-vd   & 54.39          & 74.53          & 65.08          \\
    \midrule
    RT-DETR-O~\cite{rtdetr}         & R50-vd   & 59.76          & 75.30          & 70.74          \\
    \textbf{ORDR-DETR (Ours)}       & R50-vd   & \textbf{60.97} & \textbf{76.30} & \textbf{71.89} \\
    \bottomrule
  \end{tblr}
\end{table}

% ---- Table: FAIR1M-v2.0 comparison ----
\begin{table}[!t]
  \centering
  \caption{Comparison results with state-of-the-art methods on FAIR1M-v2.0.}
  \label{tab:fair1m}
  \begin{tblr}{
      colspec = {llcc},
    }
    \toprule
    Method                          & Venue & Backbone & AP$_{50}$      \\
    \midrule
    R$^3$Det~\cite{r3det}           & AAAI2021  & R50      & 34.67          \\
    S$^2$ANet~\cite{s2anet}         & TGRS2022  & R50      & 40.07          \\
    Oriented R-CNN~\cite{orcnn}     & ICCV2021  & R50      & 43.90          \\
    PCLDet~\cite{pcldet}            & TGRS2023  & R50      & 44.45          \\
    PETDet~\cite{petdet}            & TGRS2024  & R50      & 48.61          \\
    DRNet~\cite{drnet}              & TCSVT2025 & R50      & 45.24          \\
    Deformable-DETR-O~\cite{deform} & ICLR2021  & R50      & 42.97          \\
    DINO-O~\cite{dino}              & ICLR2023  & R50      & 45.83          \\
    ARS-DETR~\cite{arsdetr}         & TGRS2024  & R50      & 42.45          \\
    OrientedFormer~\cite{oformer}   & TGRS2024  & R50      & 44.92          \\
    RHINO~\cite{rhino}              & WACV2025  & R50      & 45.13          \\
    O2-RT-DETR~\cite{o2rtdetr}      & TGRS2026  & R50-vd   & 47.97          \\
    \midrule
    RT-DETR-O~\cite{rtdetr}         & CVPR2024  & R50-vd   & 49.34          \\
    \textbf{ORDR-DETR (Ours)}       & ---       & R50-vd   & \textbf{50.97} \\
    \bottomrule
  \end{tblr}
\end{table}

Table~\ref{tab:shiprs_sota} shows that ORDR-DETR achieves 60.97 mAP on ShipRSImageNet. Compared with RT-DETR-O, it brings a gain of +1.21 mAP and gives the best overall result among the compared methods. This result further confirms the effectiveness of ORDR-DETR for fine-grained ship detection.

We further validate the effectiveness of the proposed method on the larger-scale FAIR1M-v2.0 dataset. As shown in Table~\ref{tab:fair1m}, ORDR-DETR achieves 50.97 AP$_{50}$ and brings a gain of +1.63 AP$_{50}$ over RT-DETR-O. This result shows that ORDR-DETR remains effective on more challenging large-scale benchmarks.

% ---- Ablation Studies ----
\subsection{Ablation Studies}

\subsubsection{Effectiveness of New Components}
\begin{table}[!t]
  \centering
  \caption{Ablation study of each proposed component on three datasets.}
  \label{tab:component}
  \begin{tblr}{
      cells = {c},
      cell{2}{1} = {r=3}{},
      cell{5}{1} = {r=3}{},
      cell{8}{1} = {r=3}{},
    }
    \toprule
    Dataset        & OGQR       & DQR        & mAP            & AP$_{50}$      & AP$_{75}$      \\
    \midrule
    MAR20          &            &            & 61.58          & 84.23          & 77.18          \\
                   & \checkmark &            & 62.07          & 85.06          & 78.15          \\
                   & \checkmark & \checkmark & \textbf{62.73} & \textbf{85.80} & \textbf{78.88} \\
    \midrule
    ShipRSImageNet &            &            & 59.76          & 75.30          & 70.74          \\
                   & \checkmark &            & 60.31          & 75.60          & 70.76          \\
                   & \checkmark & \checkmark & \textbf{60.97} & \textbf{76.30} & \textbf{71.89} \\
    \midrule
    FAIR1M-v2.0    &            &            & --             & 49.34          & --             \\
                   & \checkmark &            & --             & 49.82          & --             \\
                   & \checkmark & \checkmark & --             & \textbf{50.97} & --             \\
    \bottomrule
  \end{tblr}
\end{table}

We conduct ablation studies on all three datasets. Table~\ref{tab:component} shows that OGQR alone improves the baseline by +0.49 mAP on MAR20 and +0.55 mAP on ShipRSImageNet. Adding DQR further increases the gains to +1.15 mAP and +1.21 mAP, respectively. This shows that after the ranking mechanism is improved, the queries can further benefit from low-level local detail for fine-grained recognition. On FAIR1M-v2.0, OGQR and DQR bring gains of +0.48 and +1.15 AP$_{50}$, respectively, suggesting that detail refinement is more important on this dataset.

\subsubsection{Ablation on DQR}

\begin{table}[!t]
  \centering
  \caption{Ablation on the design choices of DQR on MAR20.}
  \label{tab:dqr_ablation}
  \begin{tblr}{
      colspec = {l c c c},
      cells = {c},
      column{1} = {l},
    }
    \toprule
    Variant             & Pooling             & Injection layers  & mAP            \\
    \midrule
    OGQR-only           & --                  & --                & 62.07          \\
    DQR w/o axis-aware  & global              & last two          & 62.45          \\
    DQR on early layers & axis-aware          & first two         & 62.35          \\
    \textbf{DQR (full)} & \textbf{axis-aware} & \textbf{last two} & \textbf{62.73} \\
    \bottomrule
  \end{tblr}
\end{table}

Table~\ref{tab:dqr_ablation} reports DQR design ablations on MAR20. DQR with global pooling in the last two decoder layers lifts mAP from 62.07 to 62.45, gaining +0.38 mAP. Replacing global pooling with axis-aware pooling further raises mAP to 62.73, corresponding to a gain of +0.66 mAP over the OGQR-only baseline. This suggests that oriented objects often have different structures along the long and short axes, while global pooling can weaken this directional information.

We further study where to inject the detail features. Applying DQR to the first two decoder layers gives 62.35 mAP, which is lower than the 62.73 mAP obtained when DQR is applied to the last two layers. Early decoder layers have less reliable reference boxes, so the extracted RoI features are less precise. Therefore, we use axis-aware pooling and inject DQR into the last two decoder layers.

\subsubsection{Matching Stability Analysis}

To verify the motivation of OGQR, we further analyze the matching instability during training on MAR20. Following DN-DETR~\cite{dndetr}, we compute the instability score (IS) between two consecutive epochs as the number of queries whose matched indices change. We average IS over all images to obtain the dataset-level score.

\begin{figure}[!t]
  \centering
  \includegraphics[width=\linewidth]{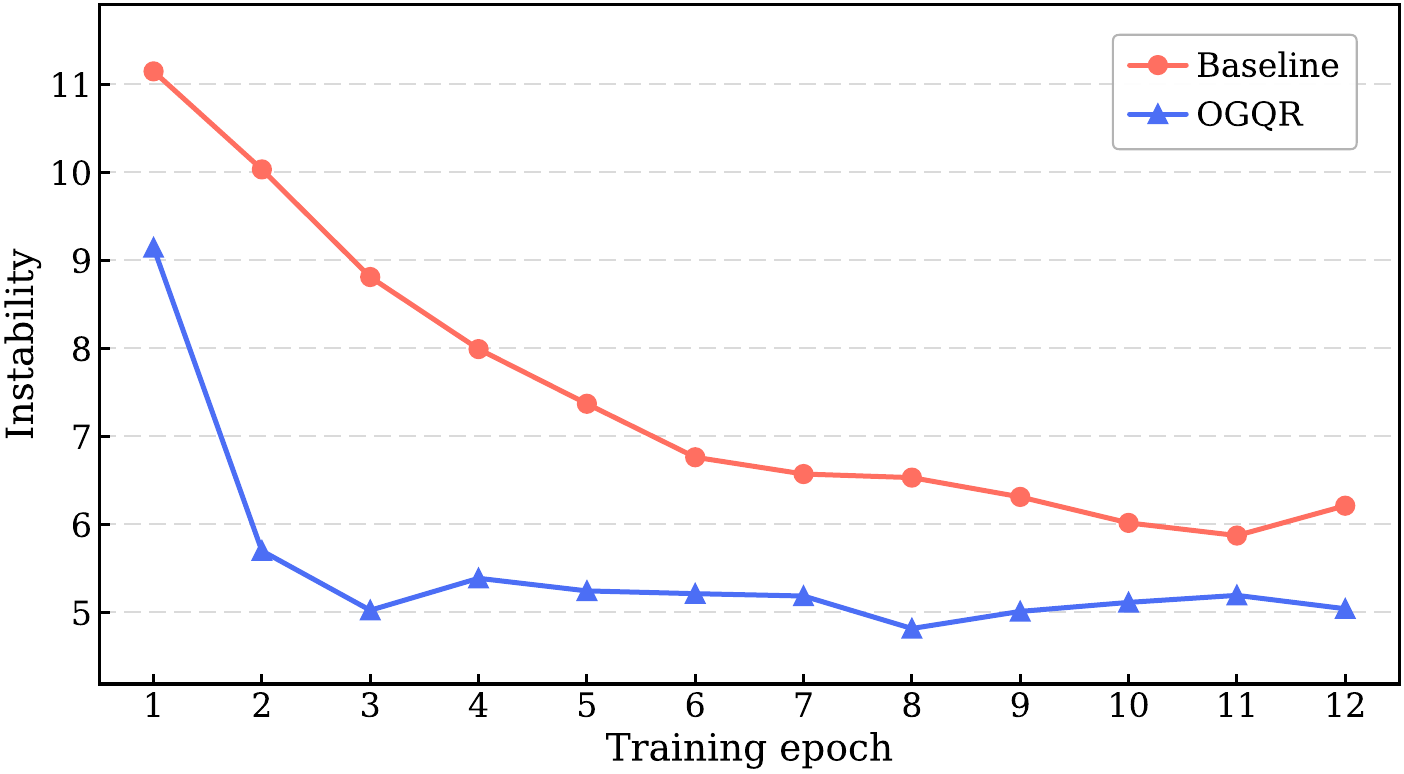}
  \caption{Comparison of matching instability during the first 12 training epochs on MAR20. IS is computed between two consecutive checkpoints, and lower values indicate more stable Hungarian matching.}
  \label{fig:is_compare}
\end{figure}

As shown in Fig.~\ref{fig:is_compare}, OGQR gives lower IS than the baseline throughout the first 12 epochs, and the gap is larger in the early stage. In FGOD, the shared classification score is easily affected by fine-grained semantic ambiguity. When this score is also used for query ranking, the matching can oscillate across adjacent epochs. After decoupling ranking from fine-grained classification, OGQR provides a cleaner ranking signal and reduces this fluctuation.

%% ===================== CONCLUSION =====================
\section{Conclusion}

We propose ORDR-DETR for fine-grained object detection in remote sensing. The main finding is that, for DETR-based FGOD, query ranking and fine-grained classification should not rely on the same score, and late decoder queries still need explicit local detail. By addressing these two issues, ORDR-DETR brings consistent gains on MAR20, ShipRSImageNet, and FAIR1M-v2.0 while preserving the end-to-end pipeline. These results suggest that better ranking signals and detail-aware query refinement are both important for DETR-based FGOD. At the same time, the current validation is still limited to oriented remote-sensing FGOD. In future work, we will extend this design to universal FGOD pipeline and broader fine-grained detection tasks.

%% ===================== REFERENCES =====================
\printbibliography

@article{fair1m,
  title   = {{FAIR1M}: A Benchmark Dataset for Fine-grained Object Recognition in High-resolution Remote Sensing Imagery},
  author  = {Sun, Xian and Wang, Peijin and Yan, Zhiyuan and Xu, Feng and Wang, Ruiping and Diao, Wenhui and Chen, Jin and Li, Jihao and Feng, Yingchao and Xu, Tao and others},
  journal = {ISPRS J. Photogramm. Remote Sens.},
  volume  = {184},
  pages   = {116--130},
  month   = {2},
  year    = {2022}
}

@article{o2rtdetr,
  title   = {Real-Time Oriented Object Detection Transformer in Remote Sensing Images},
  author  = {Ding, Zeyu and Zhao, Jiaqi and Zhou, Yong and Zhu, Hancheng and Du, Wen-Liang and Yao, Rui and El Saddik, Abdulmotaleb},
  journal = {IEEE Trans. Geosci. Remote Sens.},
  volume  = {64},
  year    = {2026},
  note    = {{A}rt. no. 5613014}
}

@inproceedings{detr,
  title     = {End-to-end object detection with transformers},
  author    = {Carion, Nicolas and Massa, Francisco and Synnaeve, Gabriel and Usunier, Nicolas and Kirillov, Alexander and Zagoruyko, Sergey},
  booktitle = {Proc. Eur. Conf. Comput. Vis. (ECCV)},
  pages     = {213--229},
  year      = {2020}
}

@inproceedings{dino,
  title     = {{DINO}: {DETR} with Improved DeNoising Anchor Boxes for End-to-End Object Detection},
  author    = {Hao Zhang and Feng Li and Shilong Liu and Lei Zhang and Hang Su and Jun Zhu and Lionel M. Ni and Heung-Yeung Shum},
  booktitle = {Proc. Int. Conf. Learn. Represent. (ICLR)},
  year      = {2023}
}

@inproceedings{rtdetr,
  title     = {{DETRs} Beat {YOLOs} on Real-Time Object Detection},
  author    = {Zhao, Yian and Lv, Wenqiang and Xu, Shilong and Wei, Jingjing and Wang, Guoqiang and Dang, Qingqing and Liu, Yi and Chen, Jie},
  booktitle = {Proc. IEEE/CVF Conf. Comput. Vis. Pattern Recognit. (CVPR)},
  pages     = {16965--16974},
  year      = {2024}
}

@inproceedings{vfl,
  title     = {{VarifocalNet}: An iou-aware dense object detector},
  author    = {Zhang, Haoyang and Wang, Ying and Dayoub, Feras and Sunderhauf, Niko},
  booktitle = {Proc. IEEE/CVF Conf. Comput. Vis. Pattern Recognit. (CVPR)},
  pages     = {8514--8523},
  year      = {2021},
  month     = {6}
}

@inproceedings{dndetr,
  title     = {{DN}-{DETR}: Accelerate {DETR} training by introducing query denoising},
  author    = {Li, Feng and Zhang, Hao and Liu, Shilong and Guo, Jian and Ni, Lionel M and Zhang, Lei},
  booktitle = {Proc. IEEE/CVF Conf. Comput. Vis. Pattern Recognit. (CVPR)},
  pages     = {13619--13627},
  year      = {2022}
}

@article{mar20,
  title   = {{MAR20}: A Benchmark for Military Aircraft Recognition in Remote Sensing Images},
  author  = {Yu, WenQi and Cheng, Gong and Wang, MeiJun and Yao, YanQing and Xie, XingXing and Yao, XiWen and Han, JunWei},
  year    = {2023},
  month   = {12},
  journal = {Nat. Remote Sens. Bull.},
  doi     = {10.11834/jrs.20222139}
}

@article{shiprs,
  title   = {{ShipRSImageNet}: A large-scale fine-grained dataset for ship detection in high-resolution optical remote sensing images},
  author  = {Zhang, Zhengning and Zhang, Lin and Wang, Yue and Feng, Pengming and He, Ran},
  journal = {IEEE J. Sel. Topics Appl. Earth Observ. Remote Sens.},
  volume  = {14},
  pages   = {8458--8472},
  year    = {2021}
}

@article{petdet,
  title   = {{PETDet}: Proposal Enhancement for Two-Stage Fine-Grained Object Detection},
  author  = {Li, Wentao and Zhao, Danpei and Yuan, Bo and Gao, Yue and Shi, Zhenwei},
  journal = {IEEE Trans. Geosci. Remote Sens.},
  volume  = {62},
  year    = {2024},
  note    = {{A}rt. no. 5602214}
}

@inproceedings{deform,
  title     = {{Deformable DETR}: Deformable transformers for end-to-end object detection},
  author    = {Zhu, Xizhou and Su, Weijie and Lu, Lewei and Li, Bin and Wang, Xiaogang and Dai, Jifeng},
  booktitle = {Proc. Int. Conf. Learn. Represent. (ICLR)},
  year      = {2021}
}

@article{arsdetr,
  title   = {{ARS}-{DETR}: Aspect Ratio-Sensitive Detection Transformer for Aerial Oriented Object Detection},
  author  = {Zeng, Ying and Chen, Yushi and Yang, Xue and Li, Qingyun and Yan, Junchi},
  journal = {IEEE Trans. Geosci. Remote Sens.},
  volume  = {62},
  year    = {2024},
  note    = {{A}rt. no. 5610315}
}

@article{oformer,
  title   = {{OrientedFormer}: An End-to-End Transformer-Based Oriented Object Detector in Remote Sensing Images},
  author  = {Zhao, Jiaqi and Ding, Zeyu and Zhou, Yong and Zhu, Hancheng and Du, Wen-Liang and Yao, Rui and El Saddik, Abdulmotaleb},
  journal = {IEEE Trans. Geosci. Remote Sens.},
  year    = {2024},
  volume  = {62},
  note    = {{A}rt. no. 5640816}
}

@inproceedings{rhino,
  title     = {Hausdorff Distance Matching with Adaptive Query Denoising for Rotated Detection Transformer},
  author    = {Lee, Hakjin and Song, MinKi and Koo, Jamyoung and Seo, Junghoon},
  booktitle = {Proc. IEEE Winter Conf. Appl. Comput. Vis. (WACV)},
  pages     = {1872--1882},
  year      = {2025},
  month     = {2}
}

@inproceedings{orcnn,
  title     = {{Oriented R-CNN} for Object Detection},
  author    = {Xie, Xingxing and Cheng, Gong and Wang, Jiabao and Yao, Xiwen and Han, Junwei},
  booktitle = {Proc. IEEE/CVF Int. Conf. Comput. Vis. (ICCV)},
  pages     = {3520--3529},
  month     = {10},
  year      = {2021}
}

@inproceedings{r3det,
  title     = {{R3Det}: Refined Single-stage Detector with Feature Refinement for Rotating Object},
  author    = {Yang, Xue and Yan, Junchi and Feng, Ziming and He, Tao},
  booktitle = {Proc. AAAI Conf. Artif. Intell.},
  volume    = {35},
  pages     = {3163--3171},
  year      = {2021}
}

@article{s2anet,
  title   = {Align Deep Features for Oriented Object Detection},
  author  = {Han, Jiaming and Ding, Jian and Li, Jie and Xia, Gui-Song},
  journal = {IEEE Trans. Geosci. Remote Sens.},
  volume  = {60},
  year    = {2022},
  note    = {{A}rt. no. 5602511}
}

@article{pcldet,
  title   = {{PCLDet}: Prototypical Contrastive Learning for Fine-Grained Object Detection in Remote Sensing Images},
  author  = {Ouyang, Lihan and Guo, Guangmiao and Fang, Leyuan and Ghamisi, Pedram and Yue, Jun},
  journal = {IEEE Trans. Geosci. Remote Sens.},
  volume  = {61},
  year    = {2023},
  note    = {{A}rt. no. 5613911}
}

@article{qedetr,
  title   = {{QEDetr}: {DETR} with Query Enhancement for Fine-Grained Object Detection},
  author  = {Dong, Chenguang and Jiang, Shan and Sun, Haijiang and Li, Jiang and Yu, Zhenglei and Wang, Jiasong and Wang, Jiacheng},
  journal = {Remote Sens.},
  volume  = {17},
  number  = {5},
  pages   = {893},
  year    = {2025}
}

@article{sfrnet,
  title   = {Fine-Grained Oriented Object Recognition via Separate Feature Refinement},
  author  = {Cheng, Gong and Li, Qingyang and Wang, Guangxing and Xie, Xingxing and Min, Lingtong and Han, Junwei},
  journal = {IEEE Trans. Geosci. Remote Sens.},
  volume  = {61},
  year    = {2023},
  note    = {{A}rt. no. 5613815}
}

@article{drnet,
  title   = {Learning Discriminative Representation for Fine-Grained Object Detection in Remote Sensing Images},
  author  = {Xie, Xingxing and Cheng, Gong and Li, Wenbo and Lang, Chunbo and Zhang, Peng and Yao, Yanqing and Han, Junwei},
  journal = {IEEE Trans. Circuits Syst. Video Technol.},
  volume  = {35},
  number  = {8},
  year    = {2025},
  month   = {8},
  pages   = {8197--8208}
}

\end{document}